\documentclass[10pt,letterpaper,twoside,twocolumn,journal,final]{IEEEtran}
\usepackage[utf8]{inputenc}
\usepackage[T1]{fontenc}
\usepackage{amssymb}
\usepackage{amsmath}
\usepackage{amsfonts}
\usepackage{nicefrac}
\usepackage{subfig}
\usepackage{graphicx}
\usepackage{microtype}
\usepackage{booktabs}
\usepackage[table]{xcolor}
\usepackage{siunitx}
\usepackage{paralist}
\usepackage{enumitem}
\usepackage[ruled,vlined,linesnumbered]{algorithm2e}
\usepackage[noadjust]{cite}
\usepackage{url}
\usepackage{scalerel}
\usepackage{tikz}
\usepackage[hidelinks]{hyperref}
\usetikzlibrary{matrix,shapes,decorations.markings,arrows,calc,svg.path}

\definecolor{orcidlogocol}{HTML}{A6CE39}
\tikzset{
    orcidlogo/.pic={
        \fill[orcidlogocol] svg{M256,128c0,70.7-57.3,128-128,128C57.3,256,0,198.7,0,128C0,57.3,57.3,0,128,0C198.7,0,256,57.3,256,128z};
        \fill[white] svg{M86.3,186.2H70.9V79.1h15.4v48.4V186.2z}
        svg{M108.9,79.1h41.6c39.6,0,57,28.3,57,53.6c0,27.5-21.5,53.6-56.8,53.6h-41.8V79.1z M124.3,172.4h24.5c34.9,0,42.9-26.5,42.9-39.7c0-21.5-13.7-39.7-43.7-39.7h-23.7V172.4z}
        svg{M88.7,56.8c0,5.5-4.5,10.1-10.1,10.1c-5.6,0-10.1-4.6-10.1-10.1c0-5.6,4.5-10.1,10.1-10.1C84.2,46.7,88.7,51.3,88.7,56.8z};
    }
}
\newcommand\orcidicon[1]{\href{https://orcid.org/#1}{\mbox{\scalerel*{
    \begin{tikzpicture}[yscale=-1,transform shape]
        \pic{orcidlogo};
    \end{tikzpicture}
}{|}}}}

\graphicspath{{figs/}}
\makeatletter
\newcommand{\removelatexerror}{\let\@latex@error\@gobble}
\def\input@path{{figs/}}
\makeatother
\DeclareMathOperator*{\argmax}{arg\,max}

\begin{document}

\title{Autoencoding with a Classifier System}

\author{Richard~J.~Preen$^{\textsuperscript{\orcidicon{0000-0003-3351-8132}}}$, Stewart~W.~Wilson$^{\textsuperscript{\orcidicon{0000-0002-7540-1031}}}$, and Larry~Bull$^{\textsuperscript{\orcidicon{0000-0002-9072-2343}}}$%
	\thanks{Manuscript received August 5, 2020; revised November 2, 2020 and February 25, 2021; accepted May 1, 2021. Date of current version May 11, 2021. \emph{(Corresponding author: Richard~J.~Preen.)}}%
	\thanks{R.~J.~Preen and L.~Bull are with the Department of Computer Science and Creative Technologies, University of the West of England, Bristol BS16 1QY, U.K. (e-mail: richard2.preen@uwe.ac.uk; larry.bull@uwe.ac.uk).}%
    \thanks{S.~W.~Wilson is with Prediction Dynamics, Concord, MA 01742 USA (e-mail: wilson@prediction-dynamics.com)}%
    \thanks{Digital Object Identifier 10.1109/TEVC.2021.3079320}%
}
 
\bstctlcite{IEEEexample:BSTcontrol}
\IEEEpubid{}
\markboth{IEEE TRANSACTIONS ON EVOLUTIONARY COMPUTATION, DOI: 10.1109/TEVC.2021.3079320}%
{PREEN \MakeLowercase{{\em et al.}}:~AUTOENCODING WITH A CLASSIFIER SYSTEM}
 
\maketitle

\begin{abstract}
Autoencoders are data-specific compression algorithms learned automatically from examples. The predominant approach has been to construct single large global models that cover the domain. However, training and evaluating models of increasing size comes at the price of additional time and computational cost. Conditional computation, sparsity, and model pruning techniques can reduce these costs while maintaining performance. Learning classifier systems (LCS) are a framework for adaptively subdividing input spaces into an ensemble of simpler local approximations that together cover the domain. LCS perform conditional computation through the use of a population of individual gating/guarding components, each associated with a local approximation. This article explores the use of an LCS to adaptively decompose the input domain into a collection of small autoencoders where local solutions of different complexity may emerge. In addition to benefits in convergence time and computational cost, it is shown possible to reduce code size as well as the resulting decoder computational cost when compared with the global model equivalent.
\end{abstract}


\section{Introduction}

\IEEEPARstart{E}{volutionary} algorithms (EAs) combined in some form with stochastic gradient descent have experienced a resurgence in their use for optimising large neural networks~\cite{Stanley:2019}. These techniques seek to construct a single large global network that covers the entire feature space. While very large networks have consistently attained state-of-the-art performance in a wide range of domains, training and evaluating networks of increasing size comes at the expense of additional time and computation. Consequently, a variety of techniques to reduce the cost while maintaining performance have been proposed, e.g., pruning and sparsity~\cite{Hua:2019,Lee:2020,Hoefler:2021}.

Dynamic neural networks~\cite{Han:2021} encompass a range of techniques to adaptively modify the structure or parameters of a neural network in response to the input and have a number of advantages in terms of accuracy and computational efficiency. Conditional computation~\cite{Bengio:2016} is one such technique that selectively activates parts of the network on a per instance basis. For example, a separate routing network has been used successfully to form layers of sparsely-gated Mixture-of-Experts in which thousands of feed-forward sub-networks may be activated~\cite{Shazeer:2017}.

Similar to most machine learning techniques, the learning classifier system (LCS) XCSF~\cite{Wilson:2001} attempts to find solutions that are accurate and maximally general over the global input space. However, it maintains the additional power to adaptively subdivide the input space into an ensemble of simpler local approximations that together cover the domain. 

XCSF performs conditional computation through the use of individual gating/guarding components associated with each local approximation. Reward is therefore allocated directly to the sub-solutions. This is in contrast with the traditional EA approach where the individual being rewarded (or reinforced) represents the overall solution to the problem, and credit is therefore much less direct in terms of rewarding the components actually responsible for the decision. Consequently, XCSF may experience faster convergence and reduced computational cost. 

Autoencoders are data-specific compression algorithms learned automatically from examples. They form a core component of many learning systems~\cite{LeCun:2015} and have significantly contributed to improvements in the current state-of-the-art for speech recognition, computer vision, and natural language processing. Autoencoders are commonly used to perform dimensionality reduction, data denoising, imputing missing data, and anomaly detection. They may be combined with a predictive component and further refined under a supervised scheme. 

Encoders and decoders of different computational complexity are required depending on the specific application. For example, a server with large amounts of available computational power can afford to spend more time encoding data, but when sent to a low power mobile device a smaller decoder may be needed. 

Usually, autoencoders are trained as single large networks using standard backpropagation techniques~\cite{Bengio:2013}. However, an ensemble of autoencoders used for image compression has recently demonstrated superior performance than \textsc{jpeg-2000}~\cite{Theis:2017}. In this approach, the most suitable autoencoder is used for compression and an additional byte then added to the coding cost in order to identify which autoencoder within the ensemble was used. 

While LCS have been extensively applied to reinforcement learning and supervised learning tasks, their use for unsupervised learning remains almost unexplored. Following on very initial work with a simple LCS~\cite{Bull:2019}, we suggest that an XCSF-like system might be capable of building an emergent ensemble of heterogeneous autoencoders with possible advantages in performance and efficiency over global model building techniques.

For example, for a specified distortion/error, a collection of LCS individuals, each smaller than the equivalent global solution, could potentially be used in a similar approach to \cite{Theis:2017}, thereby reducing byte size and reducing the cost of decoding.

\IEEEpubidadjcol

In particular, this article makes the following contributions.
\begin{enumerate}
    \item The XCSF classifier system is adapted for the autoencoder problem and tested on numerous data sets for the first time.
    \item The performance of neural networks is explored where the number of neurons as well as the connectivity are evolved, i.e., heterogeneous niched encoders may emerge.
    \item A self-adaptive scheme is introduced wherein each layer adapts to a local rate of gradient descent.
    \item A means is provided by which a target error is specified and the system automatically designs maximally compressed networks with the desired reconstruction error. This is in contrast with the traditional approach of manually specifying the network architecture and imposing predefined penalty functions and sparsity constraints.
    \item The hypothesis that LCS adaptive niching can provide improvements in performance is empirically tested by comparison with the global model building equivalent.
\end{enumerate}

The remainder of this article is organised as follows. Section~\ref{section:background} describes the XCSF classifier system, and presents an overview of the related work on neural classifiers and autoencoders. Section~\ref{section:methodology} describes the neural classifier representation and learning scheme adopted, along with the experimental method applied. Section~\ref{section:results} presents the results from experimentation on a range of publicly available data sets. Section~\ref{section:conclusion} presents our conclusions.

\section{Background}
\label{section:background}

\subsection{XCSF Classifier System}

XCSF is an accuracy-based online evolutionary machine learning system with locally approximating functions that compute classifier payoff prediction directly from the input state. XCSF can be seen as a generalisation of XCS~\cite{Wilson:1995} where the prediction is a scalar value. 

XCSF is rule-based and maintains a population of classifiers, where each classifier $cl$ consists of
\begin{inparaenum}[(i)]
    \item a condition component $cl.C$ that determines whether the rule matches input $\vec{x}$
    \item an action component $cl.A$ that selects an action $a$ to be performed for a given $\vec{x}$
    \item a prediction component $cl.P$ that computes the expected payoff for performing $a$ upon receipt of $\vec{x}$.
\end{inparaenum}%
XCSF thus generates rules of the general form: \texttt{IF} \textit{matches} $\leftarrow cl.C(\vec{x})$ \texttt{THEN} perform action $a \leftarrow cl.A(\vec{x})$ and \texttt{EXPECT} payoff $\vec{p} \leftarrow cl.P(\vec{x})$.

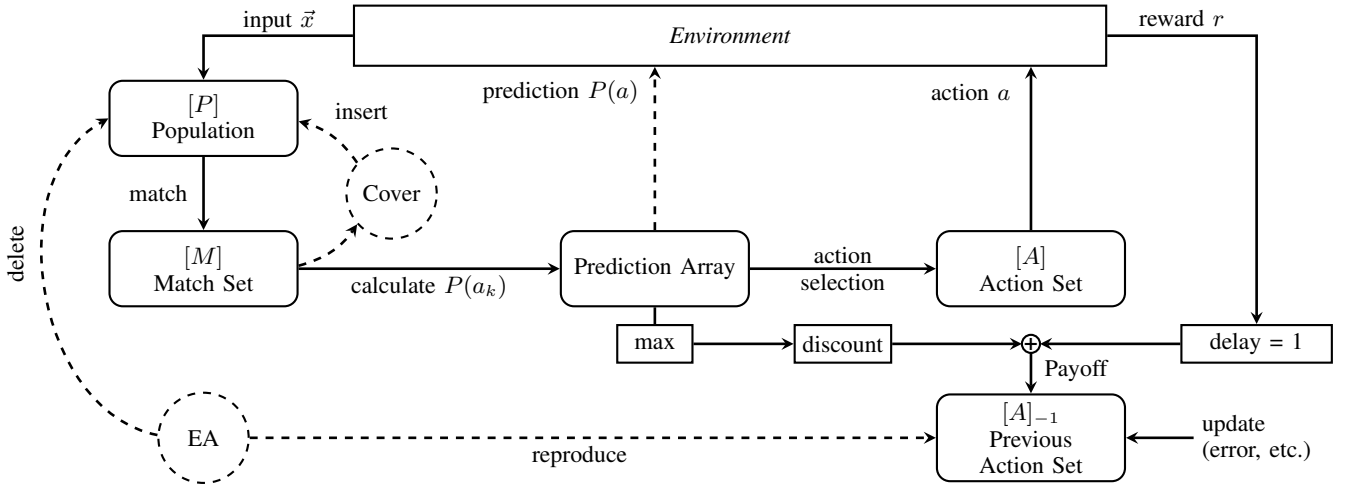
\begin{figure*}[t]
	\centering
	\small
    \begin{tikzpicture}
        % population
        \node(p) at (0,1.0)[align=center, rectangle, rounded corners=5pt, draw, thick, minimum width=25mm, minimum height=10mm] {$[P]$\\Population};
        % match set
        \node(m) at (0,-1.0)[align=center, rectangle, rounded corners=5pt, draw, thick, minimum width=25mm, minimum height=10mm] {$[M]$\\Match Set};
        % action set
        \node(a) at (11,-1.0)[align=center, rectangle, rounded corners=5pt, draw, thick, minimum width=25mm, minimum height=10mm] {$[A]$\\Action Set};
        % previous action set
        \node(preva) at (11,-3.25)[align=center, rectangle, rounded corners=5pt, draw, thick, minimum width=25mm, minimum height=10mm] {$[A]_{-1}$\\Previous\\Action Set};
        % prediction array
        \node(pa) at (6,-1.0)[align=center, rectangle, rounded corners=5pt, draw, thick, minimum width=25mm, minimum height=10mm] {Prediction Array};    
        % environment
        \node(env) at (7,2.1)[align=center, rectangle, rounded corners=0pt, draw, thick, minimum width=100mm, minimum height=8mm] {{\em Environment}};
        % delay
        \node(delay) at (14,-2.0)[align=center, rectangle, rounded corners=0pt, draw, thick, minimum width=20mm, minimum height=5mm] {{\small delay = 1}};
        % max
        \node(max) at (6,-2.0)[align=center, rectangle, rounded corners=0pt, draw, thick, minimum width=10mm, minimum height=5mm] {{\small max}};
        % discount
        \node(discount) at (8.5,-2.0)[align=center, rectangle, rounded corners=0pt, draw, thick, minimum width=10mm, minimum height=5mm] {{\small discount}};
        % dashed circles
        \node(ea) at (0,-3.25)[circle, draw, dashed, thick, minimum height=12mm] {EA};
        \node(c) at (2.5,0)[circle, draw, dashed, thick, minimum height=12mm] {Cover};
        % solid circles
        \node(plus) at (11,-2.0)[circle, draw, solid, thick, inner sep=0, outer sep=0] {{\bf \small +}};
        % line labels
        \node(payoff) at (11.6,-2.35)[] {Payoff};
        \node(calculate) at (3,-1.25)[] {calculate $P(a_k)$};
        \node(match) at (-0.6,0)[] {match};
        \node(input) at (1,2.3)[] {input $\vec{x}$};
        \node(prediction) at (10.2,1.35)[] {action $a$}; 
        \node(reinforce) at (14,-3.25)[align=left] {update\\(error, etc.)};
        \node(selection) at (8.5,-1.0)[align=center] {action\\selection}; 
        \node(reproduce) at (5,-3.5)[] {reproduce}; 
        \node(delete) at (-2.50,-0.8)[rotate=90] {delete};
        \node(insert) at (2.1,1.1)[] {insert}; 
        \node(syspred) at (4.75,1.35) [align=left] {prediction $P(a)$}; 
        \draw [black, line width=1pt]
        (pa.south) edge node [above] {} (max);
        % solid lines
        \draw [black, line width=1pt, ->, >=stealth]
        (plus) edge node [above] {} (preva)
        (p) edge node [above] {} (m)
        (m) edge node [above] {} (pa)
        (pa) edge node [above] {} (a)
        (reinforce) edge node [above] {} (preva)
        (a.north) -- (a.north|-env.south);
        \draw [black, line width=1pt, ->, >=stealth] (env) -| (p);
        \draw [black, line width=1pt, ->, >=stealth] (env) -- node[above] {reward $r$} ++(7,0) |- node[left] {} ++(0,-2) -- (delay);	
        \draw [black, line width=1pt, ->, >=stealth] (max) -- (discount);
        \draw [black, line width=1pt, ->, >=stealth] (discount) -- (plus);	
        \draw [black, line width=1pt, ->, >=stealth] (delay) -- (plus);	
        % dashed lines
        \draw [black, dashed, line width=1pt, ->, >=stealth]
        (pa.north) -- (pa.north|-env.south)
        (m) edge [bend right=20] node [above] {} (c)
        (c) edge [bend right=20] node [above] {} (p)
        (ea) edge [bend right=0] node [above] {} (preva.west)
        (ea) edge [bend left=70] node [] {} (p.west);
    \end{tikzpicture}	
    \caption{XCSF schematic illustration. For supervised learning, a single (dummy) action is performed such that $[A]=[M]$ and the system prediction is made directly accessible to the environment; classifier updates and the EA thus performed in $[M]$.}
	\label{fig:xcsf}
\end{figure*}

For each step within a learning trial, XCSF constructs a match set $[M]$ composed of classifiers in the population set $[P]$ whose $cl.C$ matches $\vec{x}$. If $[M]$ contains fewer than $\theta_{\text{mna}}$ actions, a covering mechanism generates classifiers with matching $cl.C$ and random $cl.A$. For each possible action $a_k$ in $[M]$, XCSF estimates the expected payoff by computing the fitness-weighted average as a system prediction $P(a_k)$. That is, for each action $a_k$ and classifier prediction $p_j$ in $[M]$, the system prediction $P_k = \nicefrac{\sum_j F_jp_j}{\sum_jF_j}$. A system action is then randomly or probabilistically selected during exploration, and the highest payoff action $\argmax P_k$ used during exploitation. Classifiers in $[M]$ advocating the chosen action are subsequently used to construct an action set $[A]$. The action is then performed and a scalar reward $r \in \mathbb{R}$ received, along with the next sensory input.

Upon reaching a terminal state within the environment (as is always the case in single-step problems), each classifier $cl_j \in [A]$ has its experience $exp$ incremented and fitness $F$, error $\epsilon$, and set size $as$ updated using the Widrow-Hoff delta rule with learning rate $\beta \in [0,1]$ as follows.
\begin{itemize}[label=$\triangleright$]
    \item Error: $\epsilon_j \leftarrow \epsilon_j + \beta(|r-p_j| - \epsilon_j)$
    \item Accuracy: $\kappa_j = 
        \begin{cases}
            1 & \text{if $\epsilon_j < \epsilon_0$} \\
            {\alpha(\nicefrac{\epsilon_j}{\epsilon_0})}^{-\nu} & \text{otherwise}.
        \end{cases} $ \\
        With target error threshold $\epsilon_0$ and accuracy fall-off rate $\alpha \in [0,1]$, $\nu \in \mathbb{N}_{>0}$.
    \item Relative accuracy: $\kappa_j' = \nicefrac{\kappa_j \cdot num_j}{\sum_j \kappa_j \cdot num_j}$ \\
    Where classifier numerosity initialised $num=1$.
    \item Fitness: $F_j \leftarrow F_j + \beta(\kappa_j'-F_j)$
    \item Set size estimate: $as_j \leftarrow as_j + \beta(|[A]|-as_j)$
\end{itemize}
Thereafter, $cl.C$, $cl.A$, and $cl.P$ are updated according to the representation adopted.

The EA is applied to classifiers within $[A]$ if the average set time since its previous execution exceeds $\theta_{\text{EA}}$. Upon invocation, the time stamp $ts$ of each classifier is updated. Two parents are chosen based on their fitness via roulette wheel (or tournament) selection and $\lambda$ number of offspring are created via crossover with probability $\chi$ and mutation with probability $\mu$. Offspring parameters are initialised by setting the error and fitness to the parental average, and discounted by reduction parameters for error $\epsilon_R$ and fitness $F_R$. Offspring $exp$ and $num$ are set to 1. 

If subsumption is enabled and the offspring are subsumed by either parent with sufficient accuracy ($\epsilon_j < \epsilon_0$) and experience ($exp_j > \theta_{\text{sub}}$), it is not included in $[P]$; instead the parents' micro-classifier numerosity $num$ is incremented. The resulting offspring are added to $[P]$ and the maximum (micro-classifier) population size $N$ is enforced by removing classifiers selected via roulette wheel with the deletion vote.

The deletion vote for each $cl \in [P]$ is set proportionally to the action set size estimate $as$. However, the vote is increased by a factor $\nicefrac{\overline{F}}{F_j}$ for classifiers that are sufficiently experienced ($exp_j > \theta_{\text{del}}$) and with small fitness $F_j < \delta \overline{F}$; where $\overline{F}$ is the $[P]$ mean fitness, and typically $\delta=0.1$. Classifiers selected for deletion have their (micro-classifier) $num$ decremented, and in the event that $num<1$ are removed from $[P]$.

In multi-step problems, the previous action set $[A]_{-1}$ is updated after each step with a $\gamma \in [0,1]$ discounted reward, similar to $Q$-learning, and the EA may be run therein. For regression problems, a single (dummy) action is performed such that $[A]=[M]$ and the system prediction is made directly accessible to the environment. See schematic in Fig.~\ref{fig:xcsf}.

A number of interacting pressures have been identified within XCS~\cite{Butz:2004}. A set pressure provides more frequent reproduction opportunities for more general rules. In opposition is a fitness pressure which represses the reproduction of inaccurate and over-general rules. See \cite{Bull:2015} for an overview of LCS and \cite{Butz:2006} for a detailed introduction to XCSF.

Many forms of $cl.C$, $cl.A$, and $cl.P$ have been used for classifier knowledge since the original ternary conditions, integer actions, and scalar predictions. Notable examples include, hyperrectangle~\cite{Stone:2003}, symbolic tree~\cite{Iqbal:2014}, and Haar-like feature conditions~\cite{Ebadi:2014}; least squares~\cite{Wilson:2001} and support vector predictions~\cite{Loiacono:2007}; hyperellipsoidal conditions and recursive least squares predictions~\cite{Butz:2008}; fuzzy logic~\cite{Casillas:2007}; temporally dynamic graphs~\cite{Preen:2013}, and neural networks~\cite{Bull:2002}.

Perhaps somewhat surprisingly, there had been no previous use of XCS for extracting structure within unlabelled data until the work of \cite{Tamee:2007} on clustering. They showed how the XCS generalisation mechanisms can be used to identify clusters, both their number and description.

\subsection{Evolving Neural Classifiers}

A long history of searching neural network topologies can be traced back to the origins of computing. Evolutionary, Bayesian, and reinforcement learning methods are currently widely used approaches~\cite{Stanley:2019,Elsken:2019}. A recent survey of evolutionary neural architecture search may be found in \cite{Liu:2020} wherein a large number of proposed approaches are analysed in terms of encoding and search spaces, encoding and architecture strategies, evolutionary operators and selection strategies. These approaches typically evolve large global networks whose structure and parameters remains static across different inputs and are represented as directed graphs where each node is a module representing an entire layer or sub-network, e.g., \cite{Lu:2020}.

While EAs usually combine stochastic gradient descent with an evolutionary search of the architecture, it has been suggested that EAs are competitive with stochastic gradient descent on high-dimensional problems, particularly in the case of reinforcement learning~\cite{Salimans:2017}. It has also recently been shown that evolving the network architecture without explicit weight training can produce similar results to fixed architectures where all weights are adapted~\cite{Gaier:2019}.

EAs are able to optimise neural networks even when there is no gradient information available. Moreover, several approaches exist wherein they may be combined with gradient descent techniques. Under a Lamarckian scheme, the learned weights remain as part of the genetic code for evolutionary operators to act upon~\cite{Gruau:1993}. In contrast, with Baldwinian evolution, lifetime learning is not directly reflected within the genome, but still influences selection~\cite{Hinton:1987}.

Adaptive gradient descent methods such as AdaGrad, RMSProp, and Adam have become increasingly popular. These scale the magnitude of update for each individual parameter based on various moments of the gradient. However, they frequently require some form of annealing (or warm-up schedule) to maintain early stability. These warm-up parameters typically require tuning for a specific problem and model; and the benefits over simple stochastic gradient descent with an appropriate learning rate remain controversial~\cite{Wilson:2017}.

There has also been a long history of comparison between LCS and neural networks. For example, \cite{Smith:1994} compared classifiers with the hidden neurons of a single neural network. \cite{Andersen:1993} used an EA with fitness sharing to perform layer-by-layer training of a neural network. In their approach, each individual represents a hidden neuron and the number is allowed to vary within each layer. Neurons are partitioned into sets that perform similar functions and a representative from each set is chosen to form the layer. Layers are added after a fixed number of search generations. Fitness sharing encourages the formation of different feature detectors (hidden neurons) within the population.

\cite{Bull:2002} was the first to represent LCS classifiers as neural networks: both $cl.C$ and $cl.A$ were performed within a single network rule. Self-adaptive mutation~\cite{Bull:2003} and stochastic gradient descent~\cite{OHara:2005b} were subsequently applied. In the latter, local search was performed by adapting the weights of the least fit networks in $[A]$ towards the fittest rule in the set.

\cite{OHara:2007} used neural classifiers for function approximation where gradient descent was used to update the $cl.P$ weights using the target outputs---there single networks performed $cl.C$ and $cl.P$. With the inclusion of an additional classifier network to predict the next state input, \cite{OHara:2005a} extended the approach for anticipatory LCS. More recently, \cite{Howard:2016} have explored the more biologically plausible spiking neural networks within LCS, adapting both the number of neurons and connections to perform temporal reinforcement learning.

Since the $cl.C$ is an individual component of a classifier, different knowledge representations may be paired with neural networks used for $cl.P$ or $cl.A$. For example, \cite{Lanzi:2006} used hyperrectangle $cl.C$ and neural network $cl.P$ within XCSF where the EA adapted the network topology but not the weights. This separation of concerns enable different encodings, e.g., symbolic trees or graphs, to be easily interchanged depending on the problem.

Furthermore, $cl.C$ such as hyperrectangles can be used effectively as simple gating functions. Whereas shared representations forming a single network rule~\cite{Bull:2002} enable the $cl.C$ to not only perform matching functionality, but also transform the inputs processed by $cl.P$. Cyclic graphs within XCSF have previously been used in this way to exploit the memory of processing each input, evolving gated recurrent networks~\cite{Preen:2014}.

Recently, \cite{Kim:2019} have investigated an LCS where the EA performs feature selection using bitstring conditions and a selection of convolutional neural network actions are used. LCS therefore are not mutually exclusive of novel and application-specific evolutionary NAS representations and operators, but provide a general framework within which they may be incorporated to gain additional benefits from adaptive niching and problem decomposition.

\subsection{Autoencoding}

Autoencoders are composed of an encoder and decoder, which are jointly trained to minimise the discrepancy between the original input data and its reconstruction. To capture useful structure, the encoder must be prevented from simply learning an identity function. Constraining the size of the encoder, applying regularisation techniques, penalty terms and sparsity constraints are all widely used for this purpose~\cite{Bengio:2013}.

Autoencoders have a wide range of applications even without any labelled data. For example, imputing missing data values and anomaly detection~\cite{Cao:2019}. They are frequently used in computer vision and image editing, e.g., colourising black-and-white images~\cite{Zhang:2016}, denoising images, inpainting missing regions, removing watermarks, and sharpening images~\cite{Li:2019}. Of particular use with categorical data, the trained encoder can be used to visualise data within the latent space, e.g., finding the nearest neighbours within the compressed space rather than the original input features. Multi-modal learning can be performed by jointly training an autoencoder to reconstruct multiple data modalities such as vision and language~\cite{Shi:2019}. Recurrent neural network autoencoders can be used for sequence learning~\cite{Sutskever:2014} and video compression.

A general overview of representation learning may be found in \cite{Bengio:2013}, and EA approaches to feature selection in \cite{Xue:2016}. EAs have frequently been used to design autoencoders. For example, \cite{Fernando:2016} used an EA to design the topology of compositional pattern producing networks (CPPNs) where the outputs were taken as the weights of a neural network autoencoder. The autoencoder was subsequently refined via gradient descent and the resulting gradients used to update the CPPN weights. Other examples include the EA design of denoising autoencoders for transfer learning by treating previously optimised solutions as corrupted solutions for newly encountered problems~\cite{Feng:2017}. 

Additionally, EAs have been used to design deep sparse autoencoders via differential evolution~\cite{Gong:2015} and with multiobjective optimisation~\cite{Cheng:2020}. Particle swarm optimisation has also been used to design the architecture of deep convolutional autoencoders~\cite{Sun:2019c}. Furthermore, in \cite{Feng:2019} evolutionary search is simultaneously performed on multiple tasks through the use of a denoising autoencoder. EAs have also recently been used to evolve deep neural networks for classification tasks with autoencoding components~\cite{Sun:2019b,Chen:2020}. 

Similar to training neural networks for supervised learning tasks, the predominant approach to constructing autoencoders has been to train single large global networks. However, \cite{Theis:2017} found superior compression results when compared with \textsc{jpeg-2000} by training multiple hand-crafted convolutional networks. Each autoencoder in the ensemble was optimised for a particular rate distortion trade-off. The autoencoder producing the smallest distortion was then chosen for each image and bit rate, and an extra byte added to identify the chosen autoencoder. %\cite{Reeve:2015}

Autoencoding via a single neural network has recently been used with XCS~\cite{Matsumoto:2017}. The feature inputs were initially passed through a pretrained encoder to reduce the dimensionality before performing XCS classification with an interval encoding.

\section{Methodology}
\label{section:methodology}

\begin{figure*}[t]
	\centering
	\small
	\begin{tikzpicture}
        \node(clabel) at (2,5)[] {Condition $cl.C$};
		% condition nodes
		\node(o1) at (2,4)[shape=circle, minimum size=8mm, draw] {$M$};
		\node(h1) at (1,2)[shape=circle, minimum size=8mm, draw] {$h_1$};
		\node(h2) at (3,2)[shape=circle, minimum size=8mm, draw] {$h_2$};
		\node(i1) at (0,0)[] {$x_1$};
		\node(i2) at (2,0)[] {$x_2$};
		\node(i3) at (4,0)[] {$x_3$};
        \node(mlabel) at (2,4.6)[] {if $>0.5$};
        % condition connections
		\draw [black, line width=1pt, ->, >=stealth]
		(i1) edge node [above] {} (h1)
		(i1) edge node [above] {} (h2)
		(i2) edge node [above] {} (h1)
		(i2) edge node [above] {} (h2)
		(i3) edge node [above] {} (h1)
		(i3) edge node [above] {} (h2)
		(h1) edge node [above] {} (o1)
		(h2) edge node [above] {} (o1);
		% output
		%\draw [black, line width=1pt, ->, >=stealth] (o1) edge node [above] {} ++ (0,.7);
        % condition weights
        \node(wilabel) at (0,1)[] {$w_{i,j}$};
        \node(whlabel) at (1,3)[] {$w_{j,k}$};
	\end{tikzpicture}\hspace{15mm}%
 	\begin{tikzpicture}
 	    % backprop arrow
 	    \node[label={[label distance=.5cm,text depth=-1ex,rotate=90]right:Backpropagation of error}] at (-1.25,0) {};
        \node (b1) at (-.75,4.75) {};
        \node (b2) at (-.75,0) {};
        \draw [black, dashed, line width=1pt, ->, >=stealth]
		(b1) edge node [above] {} (b2);
        \node(plabel) at (2,5)[] {Prediction $cl.P$};
		% prediction nodes
		\node(o1) at (0,4)[shape=circle, minimum size=8mm, draw] {$O_1$};
		\node(o2) at (2,4)[shape=circle, minimum size=8mm, draw] {$O_2$};
		\node(o3) at (4,4)[shape=circle, minimum size=8mm, draw] {$O_3$};
		\node(h1) at (1,2)[shape=circle, minimum size=8mm, draw] {$h_1$};
		\node(h2) at (3,2)[shape=circle, minimum size=8mm, draw] {$h_2$};
		\node(i1) at (0,0)[] {$x_1$};
		\node(i2) at (2,0)[] {$x_2$};
		\node(i3) at (4,0)[] {$x_3$};
        % prediction connections
        \draw [black, line width=1pt, ->, >=stealth]
		(i1) edge node [above] {} (h1)
		(i1) edge node [above] {} (h2)
		(i2) edge node [above] {} (h1)
		(i2) edge node [above] {} (h2)
		(i3) edge node [above] {} (h1)
		(i3) edge node [above] {} (h2)
		(h1) edge node [above] {} (o1)
		(h2) edge node [above] {} (o1)
		(h1) edge node [above] {} (o2)
		(h2) edge node [above] {} (o2)
		(h1) edge node [above] {} (o3)
		(h2) edge node [above] {} (o3);
		% outputs
		\draw [black, line width=1pt, ->, >=stealth]
		(o1) edge node [above] {} ++ (0,.7)
		(o2) edge node [above] {} ++ (0,.7)		
		(o3) edge node [above] {} ++ (0,.7);		
        % prediction weights
        \node(wilabel) at (0,1)[] {$w_{i,j}$};
        \node(whlabel) at (0,3)[] {$w_{j,k}$};
        % labels
		\node(olabel) at (5,4)[] {$\ldots O_n$};
		\node(hlabel) at (5,2)[] {$\ldots h_n$};
		\node(ilabel) at (5,0)[] {$\ldots x_n$};
	\end{tikzpicture}
	% classifier parameters
    \begin{tabular}{|c|c|c|c|c|c|c|c|}
        \toprule
        error $\epsilon$ & fitness $F$ & numerosity $num$ & experience $exp$ & set size $as$ & time stamp $ts$ & inputs seen $age$ & inputs matched $mtotal$ \\
        \bottomrule
    \end{tabular}
    \caption{Neural classifier knowledge representation. Separate fully-connected feed-forward networks calculate classifier matching and prediction. Each layer is encoded as a vector of weights (and biases), along with a binary vector indicating whether each connection is active, an activation function, and gradient descent rate. Each layer maintains its own vector of mutation rates $\vec{\mu}$.}
	\label{fig:classifier}
\end{figure*}
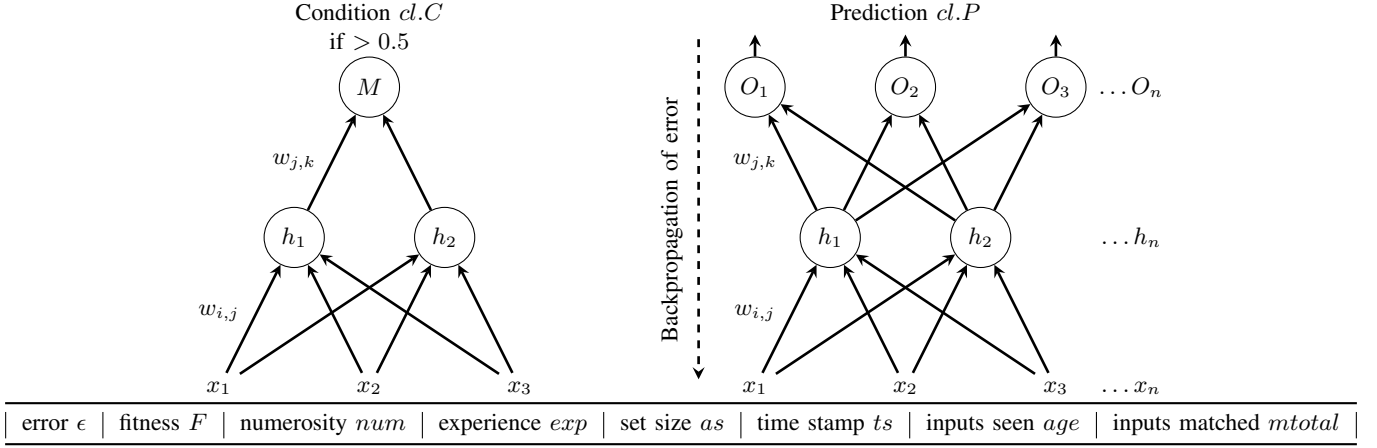

Here, we use a derivative of XCSF to explore the effects of adaptive niching in the automatic design of an ensemble of multi-layer perceptron autoencoders. That is, each classifier is trained to reproduce its inputs via a much smaller (encoding) hidden layer. Each $cl.C$ and $cl.P$ is a separate fully-connected neural network, as illustrated in Fig.~\ref{fig:classifier}. Each network is composed of hidden scaled exponential linear units (SELUs)~\cite{Klambauer:2017}, and logistic outputs. The $cl.C$ output layer contains a single neuron that determines whether the rule matches a given input. The $cl.P$ (decoding) output layer contains as many output neurons as inputs.

A population of $N=500$ classifiers are initialised randomly and undertake Lamarckian learning. That is, after the application of evolutionary operators to both $cl.C$ and $cl.P$ during reproduction, stochastic gradient descent updates $cl.P$ during reinforcement. The resulting $cl.P$ weights are copied to offspring upon parental selection.

During instantiation of $[P]$ the weights of each network are initialised with small random values sampled from a Gaussian normal distribution with mean $m=0$ and standard deviation $\sigma=0.1$. Biases are zero initialised. Should covering be triggered at any stage, networks with random weights and biases are generated by the same method until the network matches the current input, however using a larger $\sigma=1$. Upon receipt of $\vec{x}$, $[M]$ is formed by adding all $cl \in [P]$ whose $cl.C$ outputs a value greater than 0.5.

Classifier reinforcement and the EA take place within $[M]$. The $[M]$ fitness-weighted average prediction is also used for system output as is usual in XCSF. However, here learning consists of updating the matching error, which is derived from the mean squared error (MSE) with respect to $\vec{x}$ and the corresponding values on each output neuron $\vec{O}$ of a rule in the current $[M]$ using the modified Widrow-Hoff delta rule with learning rate $\beta$:
\begin{equation}
\epsilon_j \leftarrow \epsilon_j + \beta \biggl[ \frac{1}{n}\sum_{i=1}^{n}{{(x_i - O_i)}^2} - \epsilon_j \biggr]
\end{equation}
Subsequently, each $cl.P$ within $[M]$ is updated using simple stochastic gradient descent~\cite{Rumelhart:1986} with a layer-specific learning rate $\eta \in \mathbb{R}_{>0}$ and momentum $\omega \in [0,1]$. That is, the chain rule is applied at match time $t$ to compute the partial derivative of the error with respect to each weight $\nicefrac{\partial \mathcal{E}}{\partial w}$, and the weight change:
\begin{equation}
\Delta w_t = -\eta \nicefrac{\partial \mathcal{E}}{\partial w_t} + \omega \Delta w_{t-1}
\end{equation}
Gradient descent is not applied to $cl.C$. 

Following~\cite{Bull:2003} crossover is omitted and self-adaptive mutation used. However, here each layer within each classifier maintains a vector of mutation rates initially seeded randomly from a uniform distribution $\vec{\mu} \sim U[\mu_{\text{min}},1]$. These parameters are passed from parent to offspring. The offspring then applies each of these mutation rates to itself using a Gaussian distribution, i.e., $\mu_i' = \mu_i e^{\mathcal{N}(0,1)}$, before mutating the rest of the rule at the resulting rate. This is similar to the approach used in evolution strategies (ES) where the mutation rate is a locally evolving entity in itself, i.e., it adapts during the search process. Self-adaptive mutation not only reduces the number of hand-tunable parameters of the EA, it has also been shown to improve performance. Here, four types of mutation are explored such that for each layer:

\begin{itemize}
    \item Weights and biases are adapted through the use of a single self-adaptive mutation rate, which controls the $\sigma$ of a random Gaussian added to each weight and bias. This is also similar to the approach used in ES.
    \item A second self-adaptive rate controls the number of hidden neurons to add or remove. This value is discretised into the range [-$h_M$,$h_M$] with $h_M$ determining the maximum number of neurons that may be added or removed per mutation event. Pressure to evolve minimally sized networks is achieved by altering the population size enforcement mechanism as follows. Each time a classifier must be removed, two classifiers are selected via roulette wheel with the deletion vote as described above and then the rule with the most hidden layer nodes is deleted.
    \item To adapt the rate of gradient descent, each layer maintains its own $\eta$. These values are constrained [$10^{-4},0.01$] and seeded uniformly random. A third self-adaptive mutation rate controls the $\sigma$ of a random Gaussian added to each $\eta$, similar to weight adaptation. \cite{Wyatt:2005} have previously shown how the self-adaptation of local search parameters can speed learning within XCS. Here it enables the learning rate to continually adapt to the parameters throughout the search process, e.g., potentially performing smaller updates for parameters associated with frequently occurring features, and larger updates for parameters associated with infrequent features.
    \item A fourth self-adaptive rate controls the probability of enabling or disabling each connection within the layer. This may encourage a more efficient sparse representation within the networks. Networks are always initialised fully-connected. When a connection is disabled, the corresponding weight value is set to zero and is excluded from mutation and gradient descent updates. Upon activation, the weight is set to a small random value $\sim \mathcal{N}(0,0.1)$. When connection mutation is enabled, the connections of newly added neurons are activated with 50\% probability. See outline in Algorithm~\ref{alg}.
\end{itemize}

\begin{figure}[t]
    \removelatexerror
    \begin{algorithm}[H]
    	\SetNoFillComment
    	\small
    	\DontPrintSemicolon
    	$\{c1p, c2p\} \in [M] \leftarrow SelectParentsRouletteFitness()$\;
    	\For{$\nicefrac{\lambda}{2}$ number of times} {
    		$\{c1, c2\} \leftarrow Copy(c1p, c2p)$\;
    	    \For{$cl \in \{c1, c2\}$} {
    	        \For{layer $\ell \in \{cl.C, cl.P\}$} {
    	            \tcp{self-adapt mutation rates}
    	            $\ell.\mu_i \leftarrow \ell.\mu_i e^{\mathcal{N}(0,1)}$\;
    	            \tcp{mutate gradient descent rate}
     	            $\ell.\eta \leftarrow \ell.\eta + \mathcal{N}(0,\ell.\mu_1)$\;
     	            \tcp{mutate number of neurons}
     	            $\ell.h \leftarrow \ell.h + round((2 \times \ell.\mu_2 - 1) \times h_M)$\;
     	            \tcp{mutate connectivity}
    	            \For{$w_i \in \ell.\vec{w}$} {
    	                \If{$U[0,1] < \ell.\mu_3$} {
    	                    \If{$a_i$ is enabled} {
    	                        $w_i, a_i \leftarrow 0$, disabled\;
    	                    } \Else {
    	                        $w_i, a_i \leftarrow \mathcal{N}(0,0.1)$, enabled\;
    	                    }
    	                }
    	            }
    	            \tcp{mutate weight magnitudes}
    	            \For{$w_i \in \ell.\vec{w}$} {
    	                \If{$a_i$ is enabled} {
    	                    $w_i \leftarrow w_i + \mathcal{N}(0,\ell.\mu_4)$\;
    	               }
    	            }
    	        }
            }
        }  
    	\caption{Offspring Creation.}
    	\label{alg}
    \end{algorithm}
\end{figure}

\begin{table}[t]
	\caption{Learning Parameters}
    \centering
    \begin{tabular}{p{54mm}cc}
        \toprule
        Description & Parameter & Value \\
        \midrule
        Maximum population size (in micro-classifiers) & $N$ & 500 \\
        Population initialised with random classifiers & $P_{\text{init}}$ & true \\
        Target error, under which accuracy is set to 1 & $\epsilon_0$ & 0.01 \\
        Update rate for fitness, error, and set size & $\beta$ & 0.1 \\
        Accuracy offset (1=disabled) & $\alpha$ & 1 \\
        Accuracy slope & $\nu$ & 10 \\
        Fraction of classifiers to increase deletion vote & $\delta$ & 0.1 \\
        Classifier deletion threshold & $\theta_{\text{del}}$ & 20 \\
        Classifier initial fitness & $F_I$ & 0.01 \\
        Classifier initial error & $\epsilon_I$ & 0 \\
        Offspring fitness reduction (1=disabled) & $F_R$ & 0.1 \\
        Offspring error reduction (1=disabled) & $\epsilon_R$ & 1 \\ 
        Minimum number of actions in $[M]$ & $\theta_{\text{mna}}$ & 1 \\
        EA invocation frequency & $\theta_{\text{EA}}$ & 50 \\
        Number of offspring per EA invocation & $\lambda$ & 2 \\
        Crossover probability & $\chi$ & 0 \\
        Minimum self-adaptive mutation value & $\mu_{\text{min}}$ & $10^{-4}$ \\
        Stochastic gradient descent momentum & $\omega$ & 0.9 \\
        Initial number of hidden neurons & $h_I$ & 1 \\
        Max.\ neurons added or removed per mut.\ & $h_M$ & \{1,2,5\} \\
        Whether EA subsumption is performed & \em{EASubsume} & false \\
        Whether set subsumption is performed & \em{SetSubsume} & false \\
        \bottomrule
    \end{tabular}
    \label{table:params}
\end{table}

Since the possibility exists that a $cl.C$ (not generated through covering) may never match any inputs, any classifiers that have not matched any inputs within 10000 trials since creation are selected for removal during population deletion. 

As a measure of generalisation we report the fraction of inputs matched by the single best rule $cl^{*}_{\text{mfrac}}$. This rule is determined as follows. If no classifier has an error below $\epsilon_0$, the classifier with the lowest error is chosen. If more than one classifier has an error below $\epsilon_0$, the classifier that matches the largest number of inputs is used. 

To test the hypothesis that LCS adaptive niching can improve performance, we compare results with the same system, where however $cl.C$ always match $\vec{x}$. Classifier updates and the EA are thus performed within $[P]$, and single networks that cover the entire state-space are designed. This configuration operates as a more traditional EA, and we henceforth refer to this system as the EA. The EA acts as a control with which we can make direct comparison since it operates with the same evolutionary and gradient descent operators, and uses the same parameters. Any future changes or improvements made to the EA may be incorporated within XCSF.

When comparing XCSF and the EA on a single data set, we use the Wilcoxon ranked-sum test, with the null hypothesis that all observed results come from the same distribution. To measure the performance across the first 100000 trials, we also present the area under the curve (AUC) results using a composite Simpson's rule applied to the mean errors.

When making comparisons across multiple data sets we follow the recommendations of \cite{Demsar:2006} and use the Wilcoxon signed-rank test with the null hypothesis that taken across all data sets there is no difference in performance. All tests are applied with a 95\% confidence interval.

The following publicly available data sets are used for initial evaluation from \url{https://www.openml.org}
\begin{enumerate}
    \item \textsc{usps digits}: 256 features; 10 classes; 9298 instances. OpenML ID: 41082.
    \item \textsc{mnist digits}: 784 features; 10 classes; 70000 instances. OpenML ID: 554.
    \item \textsc{mnist fashion}: 784 features; 10 classes; 70000 instances. OpenML ID: 40996.
    \item \textsc{cifar10}: 3072 features; 10 classes; 60000 instances. OpenML ID: 40927.
\end{enumerate}

Table~\ref{table:params} lists the parameters used. Most values are typical defaults (e.g., \cite{Butz:2008}), with the primary exception of a larger $\nu$ increasing fitness pressure. This was found to speed convergence of XCSF and the EA similarly during initial testing. All graphs presented depict mean $[P]$ values over 10 runs for 100000 trials. Inputs are scaled $[0,1]$ and instances are drawn at random. 90\% of the sample instances are used for training and 10\% reserved for testing. 

\section{Results}
\label{section:results}

\subsection{Neuron Growth Rates}

\begin{figure*}[t]
    \subfloat[\textsc{mnist digits} (784 inputs).]{%
        \includegraphics[width=\textwidth]{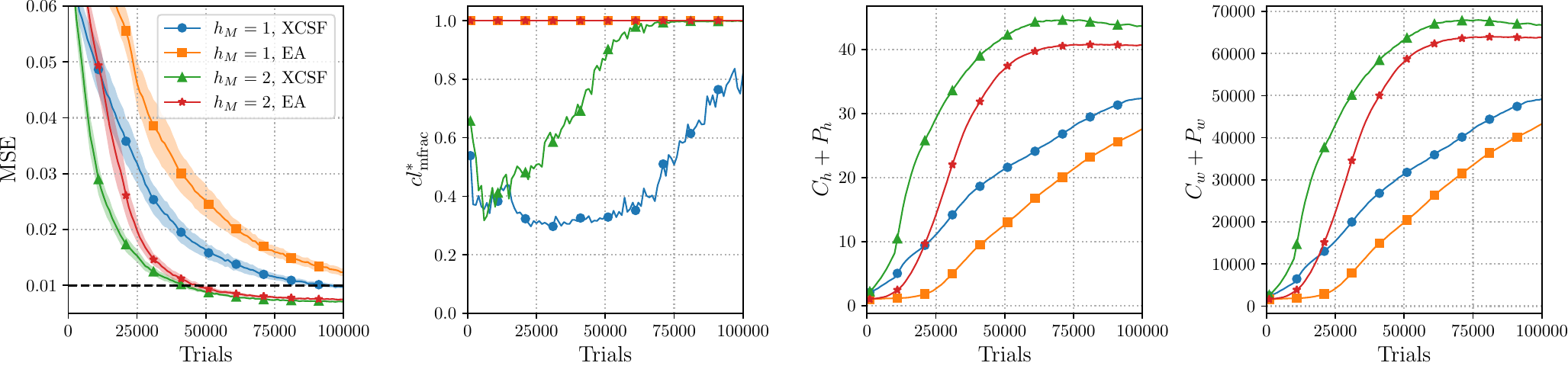}}
    \hfill
    \subfloat[\textsc{mnist fashion} (784 inputs).]{%
        \includegraphics[width=\textwidth]{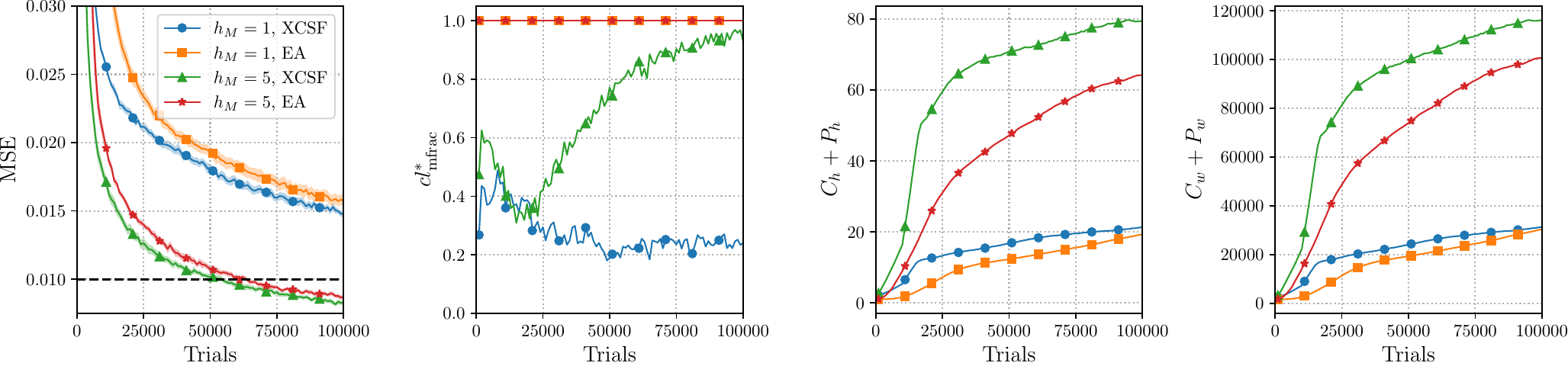}}
    \caption{The effect of maximum growth rates on the \textsc{mnist} data sets. Shown are the mean squared error (MSE), fraction of inputs matched by the best rule ($cl^{*}_{\text{mfrac}}$), condition hidden neurons ($C_h$), and prediction hidden neurons ($P_h$) for the EA and XCSF with different maximum neuron growth/removal per mutation event $h_M$.}
    \label{fig:growth}
\end{figure*}

The performance of XCSF and the EA without connection mutation on the \textsc{mnist} data sets is shown in Fig.~\ref{fig:growth}. As can be seen, across the 100000 trials with a maximum growth rate of $h_M=1$, XCSF achieves a smaller error than the EA. On both \textsc{mnist digits} and \textsc{mnist fashion}, XCSF has a smaller AUC (2287.95, 2004.95) than the EA (3184.24, 2339.69).

From Table~\ref{table:errors} it can be seen that when comparing early learning performance at 20000 trials on \textsc{mnist digits} with $h_M=1$, the XCSF mean error is significantly smaller than the EA. Similarly on \textsc{mnist fashion} after 20000 trials, the XCSF mean error is significantly smaller than the EA. 

\begin{table}[t]
    \centering
	\caption{Autoencoding errors. Statistically significant values in boldface. Shaded rows with smallest mean. Trials in 1000s. Maximum neuron growth rate ($h_M$); maximum number of neurons ($h_{\text{max}}$); with connection mutation enabled (cmut).}
    \begin{tabular}{p{18mm}rcccc}
        \toprule
		Algorithm & Trials & $h_M$ & MSE $\pm$ SE & Min & Median \\
        \midrule
        \underline{\textsc{mnist digits}} &&&&&\\
        \rowcolor[HTML]{EFEFEF}
        XCSF & 20 & 1 & \textbf{0.0373$\pm$0.0043} & 0.0218 & 0.0316\\
        EA & 20 & 1 & 0.0570$\pm$0.0036 & 0.0307 & 0.0619\\
        \rowcolor[HTML]{EFEFEF}
        XCSF & 97 & 1 & \textbf{0.0097$\pm$0.0003} & 0.0082 & 0.0098\\
        EA & 97 & 1 & 0.0128$\pm$0.0036 & 0.0097 & 0.0126\\
        \rowcolor[HTML]{EFEFEF}
        XCSF & 20 & 2 & \textbf{0.0182$\pm$0.0015} & 0.0128 & 0.0174\\
        EA & 20 & 2 & 0.0285$\pm$0.0044 & 0.0150 & 0.0259\\ 
        \rowcolor[HTML]{EFEFEF}
        XCSF & 43 & 2 & 0.0098$\pm$0.0004 & 0.0080 & 0.0101\\
        EA & 43 & 2 & 0.0108$\pm$0.0006 & 0.0083 & 0.0106\\
        \\
        \underline{\textsc{mnist fashion}} &&&&&\\
        \rowcolor[HTML]{EFEFEF}
        XCSF & 20 & 1 & \textbf{0.0222$\pm$0.0003} & 0.0211 & 0.0221\\
        EA & 20 & 1 & 0.0252$\pm$0.0006 & 0.0220 & 0.0251\\   
        \rowcolor[HTML]{EFEFEF}
        XCSF & 100 & 1 & \textbf{0.0148$\pm$0.0002} & 0.0136 & 0.0147\\
        EA & 100 & 1 & 0.0157$\pm$0.0004 & 0.0133 & 0.0156\\    
        \rowcolor[HTML]{EFEFEF}
        XCSF & 20 & 5 & \textbf{0.0137$\pm$0.0004} & 0.0119 & 0.0134\\
        EA & 20 & 5 & 0.0149$\pm$0.0002 & 0.0140 & 0.0149\\
        \rowcolor[HTML]{EFEFEF}
        XCSF & 55 & 5 & \textbf{0.0099$\pm$0.0002} & 0.0091 & 0.0099\\
        EA & 55 & 5 & 0.0105$\pm$0.0002 & 0.0099 & 0.0103\\
        \\
        \underline{\textsc{usps digits}} &&&&&\\
        \rowcolor[HTML]{EFEFEF}
        XCSF & 16 & 1 & \textbf{0.0097$\pm$0.0005} & 0.0071 & 0.0092\\
        EA & 16 & 1 & 0.0192$\pm$0.0046 & 0.0091 & 0.0112\\
        \rowcolor[HTML]{EFEFEF}
        XCSF-cmut & 16 & 1 & \textbf{0.0115$\pm$0.0006} & 0.0069 & 0.0120\\
        EA-cmut & 16 & 1 & 0.0327$\pm$0.0038 & 0.0147 & 0.0354\\
        \rowcolor[HTML]{EFEFEF}
        XCSF $h_{\text{max}}$=12 & 100 & 1 & \textbf{0.0108$\pm$0.0002} & 0.0099 & 0.0107\\
        EA $h_{\text{max}}$=12 & 100 & 1 & 0.0127$\pm$0.0002 & 0.0115 & 0.0128\\       
        \\
        \underline{\textsc{cifar10}} &&&&&\\
        \rowcolor[HTML]{EFEFEF}
        XCSF & 20 & 5 & \textbf{0.0406$\pm$0.0044} & 0.0240 & 0.0374\\
        EA & 20 & 5 & 0.0591$\pm$0.0007 & 0.0546 & 0.0595\\
        \rowcolor[HTML]{EFEFEF}
        XCSF-cmut & 20 & 5 & \textbf{0.0179$\pm$0.0012} & 0.0134 & 0.0171\\
        EA-cmut & 20 & 5 & 0.0302$\pm$0.0036 & 0.0184 & 0.0265\\     
        \rowcolor[HTML]{EFEFEF}
        XCSF & 100 & 5 & \textbf{0.0135$\pm$0.0006} & 0.0107 & 0.0140\\
        EA & 100 & 5 & 0.0163$\pm$0.0009 & 0.0131 & 0.0151\\
        \rowcolor[HTML]{EFEFEF}
        XCSF-cmut & 100 & 5 & 0.0099$\pm$0.0002 & 0.0090 & 0.0099\\
        EA-cmut & 100 & 5 & 0.0101$\pm$0.0002 & 0.0094 & 0.0102\\
        \bottomrule
    \end{tabular}
    \label{table:errors}
\end{table}

On \textsc{mnist digits} with $h_M=1$, the mean XCSF error reaches $\epsilon_0$ after 97000 trials, whereas the EA does not do so within the 100000 trials run. Comparing performance at 97000 trials, the XCSF mean error is significantly smaller than the EA. While neither XCSF nor the EA with $h_M=1$ were able to reach the target error within 100000 trials on \textsc{mnist fashion}, XCSF has a significantly smaller error after 100000 trials than the EA.

Increasing the growth rates clearly results in faster error convergence for both the EA and XCSF, with significantly smaller errors observed when compared with $h_M=1$ after 100000 trials. On \textsc{mnist digits} with $h_M=2$, the mean XCSF error reaches $\epsilon_0$ after 43000 trials, and the EA does so after 47000 trials. The $h_M=2$ XCSF AUC = 1415.97 and EA AUC = 1801.79, showing again that XCSF is faster than the EA and that $h_M=2$ results in a smaller error across the whole 100000 trials. 

\begin{figure}[t]
    \begin{tikzpicture}
        \node[label={[text depth=-1ex,rotate=90]right:Trials}] at (-0.4,1.25) {};
        \node (b1) at (0,3.5) {};
        \node (b2) at (0,0) {};
        \draw [black, dashed, line width=1pt, ->, >=stealth]
		(b1) edge node [above] {} (b2);
    \end{tikzpicture}
    \subfloat[XCSF]{\includegraphics[width=0.42\columnwidth]{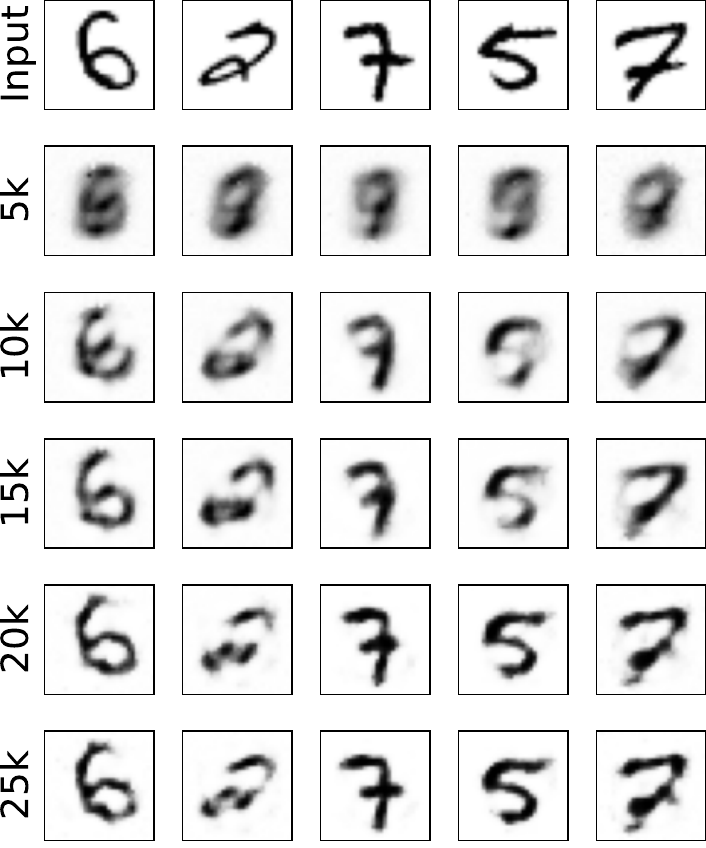}}
    \hfill
    \subfloat[EA]{\includegraphics[width=0.42\columnwidth]{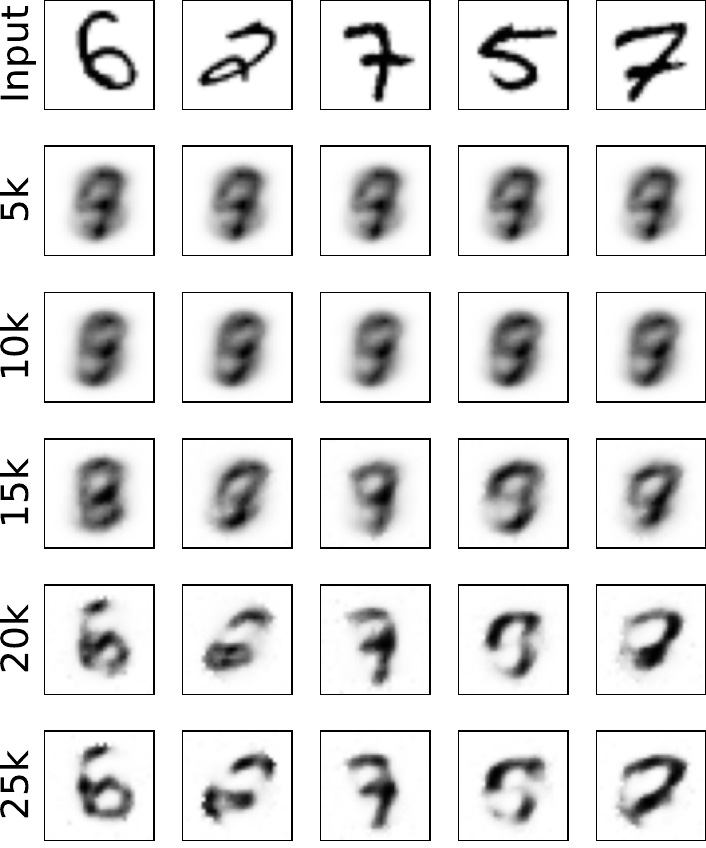}}
    \caption{\textsc{mnist digits} test set reconstruction. Single run with of the EA and XCSF for 25000 trials; no connection mutation; $h_M=2$. XCSF MSE = 0.0122 and $P_h=28.2$, $cl^{*}_{\text{mfrac}}=1$; EA MSE = 0.0253 and $P_h=6.1$.}
    \label{fig:digits_recon}
\end{figure}

XCSF early convergence with $h_M=2$ on \textsc{mnist digits} is again faster than the EA after 20000 trials. This difference in early learning performance can be observed qualitatively in Fig.~\ref{fig:digits_recon}, which shows the XCSF and EA reconstructions of a sample of images from the \textsc{mnist digits} test set over the first 25000 trials.

While the mean XCSF error after 100000 trials with $h_M=2$ on \textsc{mnist digits} is not significantly different than the EA, the XCSF mean, min and median are all smaller. 

With $h_M=5$ on \textsc{mnist fashion}, XCSF has a smaller AUC (1212.36) than the EA (1381.6) and early convergence is again faster with XCSF. After 20000 trials, XCSF has a significantly smaller error than the EA. Furthermore, XCSF reaches $\epsilon_0$ after only 55000 trials, whereas the EA reaches the threshold after 64000 trials. Comparing performance after 55000 trials, shows that XCSF has a significantly smaller error.

\begin{figure}[t]
    \includegraphics[width=\columnwidth]{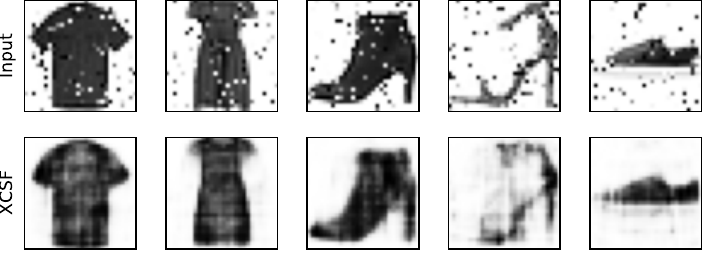}
    \caption{XCSF reconstruction of samples from \textsc{mnist fashion} test set with 10\% salt and pepper noise after 100000 trials. Training $\text{MSE}=0.0082$ and $P_h=67.2$.}
    \label{fig:fashion_denoise_10}
\end{figure}

\begin{figure}[t]
    \includegraphics[width=\columnwidth]{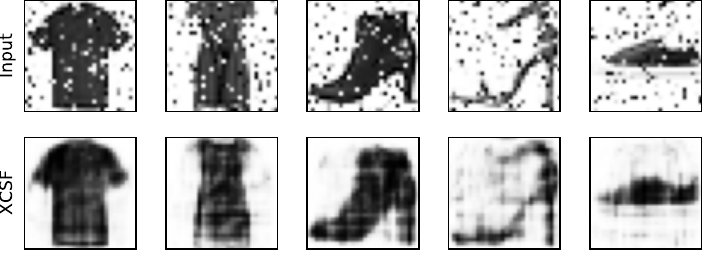}
    \caption{XCSF reconstruction of samples from \textsc{mnist fashion} test set with 20\% salt and pepper noise after 100000 trials. Training $\text{MSE}=0.0082$ and $P_h=67.2$.}
    \label{fig:fashion_denoise_20}
\end{figure}

\begin{figure}[t]
    \includegraphics[width=\columnwidth]{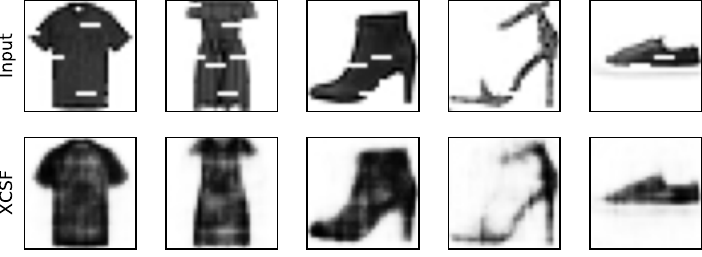}
    \caption{XCSF reconstruction of samples from \textsc{mnist fashion} test set with random cut out after 100000 trials. Training $\text{MSE}=0.0082$ and $P_h=67.2$.}
    \label{fig:fashion_cutout}
\end{figure}

Fig.~\ref{fig:fashion_denoise_10} and Fig.~\ref{fig:fashion_denoise_20} show the XCSF reconstruction after 100000 trials on \textsc{mnist fashion} where 10\% and 20\% salt and pepper noise has been added to the test images presented as input. As can be seen, an efficient representation has been learned, which can be used to effectively denoise the data. Fig.~\ref{fig:fashion_cutout} shows the XCSF reconstruction where random cutout has been applied to the test images, showing how the learned representation can be used to impute missing values.

\subsection{Feature Selection}

\begin{figure*}[t]
    \subfloat[\textsc{usps digits} (256 inputs).]{%
        \includegraphics[width=\textwidth]{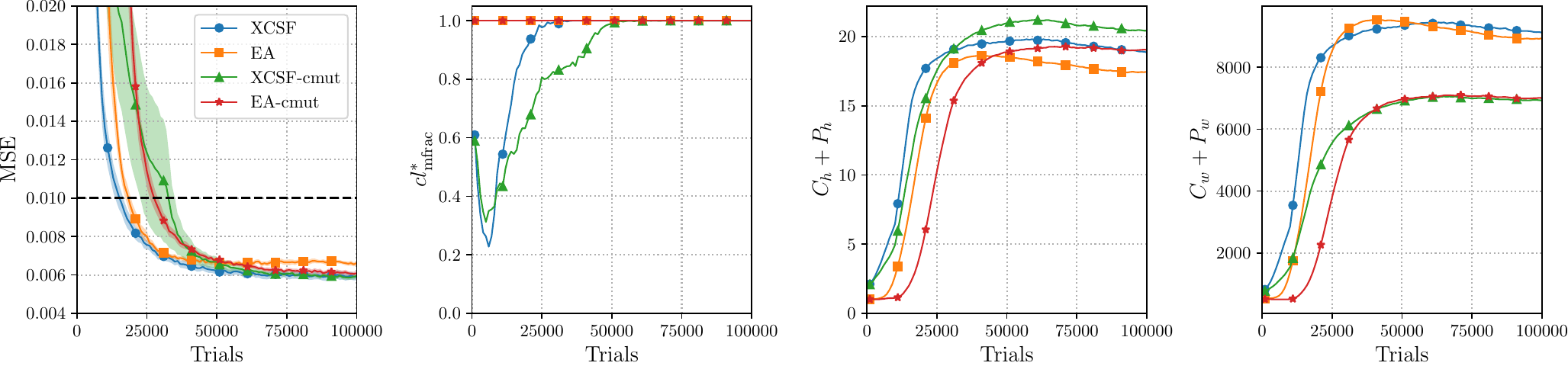}}
    \hfill
    \subfloat[\textsc{cifar10} (3072 inputs).]{%
        \includegraphics[width=\textwidth]{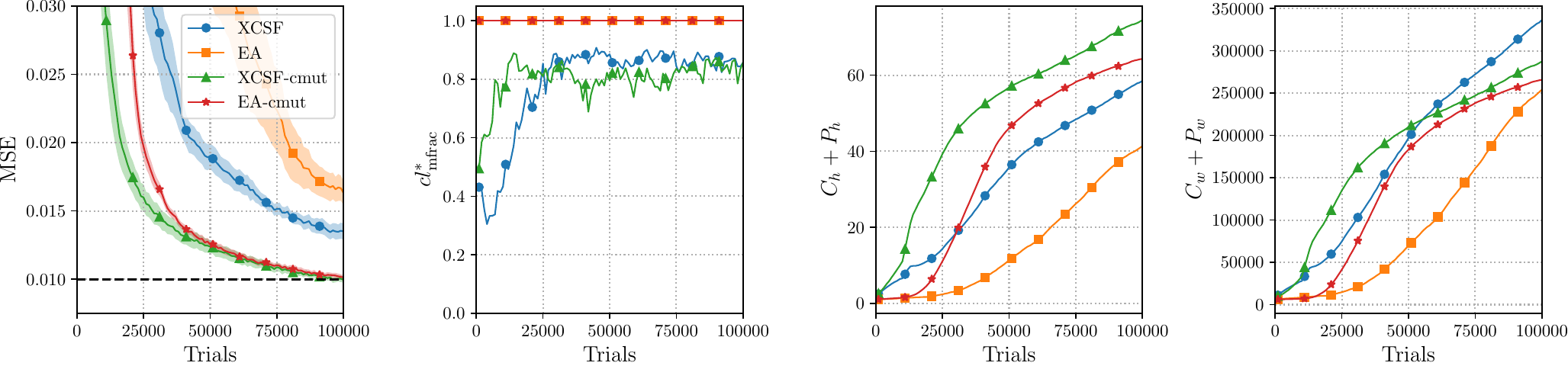}}
    \caption{The effect of feature selection on \textsc{usps digits} and \textsc{cifar10}. Shown are the mean squared error (MSE), fraction of inputs matched by the best rule ($cl^{*}_{\text{mfrac}}$), total number of condition ($C_h$) and prediction hidden neurons ($P_h$), and the total number of non-zero condition ($C_w$) and prediction weights ($P_w$) for the EA with (star) and without connection mutation (square), as well as XCSF with (triangle) and without connection mutation (circle). For \textsc{usps}, $h_M=1$ and for \textsc{cifar10}, $h_M=5$.}
    \label{fig:selection}
\end{figure*}

The performance of XCSF and the EA with and without connection mutation on \textsc{usps digits} and \textsc{cifar10} is shown in Fig.~\ref{fig:selection} and a summary of error scores in Table~\ref{table:errors}. As can be seen, across the 100000 trials, XCSF achieves a smaller error than the EA. On both data sets without connection mutation, XCSF has a smaller AUC (893.56, 2568.85) than the EA (1108.87, 3722.06). Similarly with connection mutation enabled, XCSF has a smaller AUC (1148.17, 1607.72) than the EA (1439.88, 2053.15). While the AUCs with connection mutation are larger on \textsc{usps digits}, they are smaller on \textsc{cifar10}, showing that connection mutation can be beneficial to learning. Moreover, with connection mutation enabled the number of neurons grows to a larger number, whilst the number of non-zero weights is smaller, showing that a more sparse representation is learned.

Without connection mutation on \textsc{usps digits}, XCSF with $h_M=1$ reaches $\epsilon_0$ after 16000 trials, compared with the EA which reaches the threshold after 19000 trials. When connection mutation is enabled, XCSF reaches $\epsilon_0$ after 33000 trials, and the EA after 28000 trials. However, comparing performance after 16000 trials (i.e., when $\epsilon_0$ is reached without connection mutation), the XCSF error without connection mutation is not significantly different than with connection mutation enabled.

Early MSE on \textsc{usps digits} with XCSF both with and without connection mutation is significantly smaller than the EA. For example, after 16000 trials without connection mutation, the XCSF error is significantly smaller than the EA. Similarly at the same number of trials with connection mutation, the XCSF error is significantly smaller than the EA.

On \textsc{cifar10} with $h_M=5$, neither XCSF nor the EA reach $\epsilon_0$ after 100000 trials without connection mutation. When connection mutation is used, XCSF reaches $\epsilon_0$ after 100000 trials, whereas the EA does not. Comparing errors after 100000 trials shows that XCSF without connection mutation has a significantly larger error than XCSF with connection mutation. Furthermore, XCSF without connection mutation has a significantly smaller error than the EA without connection mutation. Finally, while there is no significant difference when comparing XCSF and the EA with connection mutation, the XCSF mean, min and median are all smaller.

\subsection{Heterogeneous Niched Ensembles}

\begin{figure*}[t]
    \includegraphics[width=\textwidth]{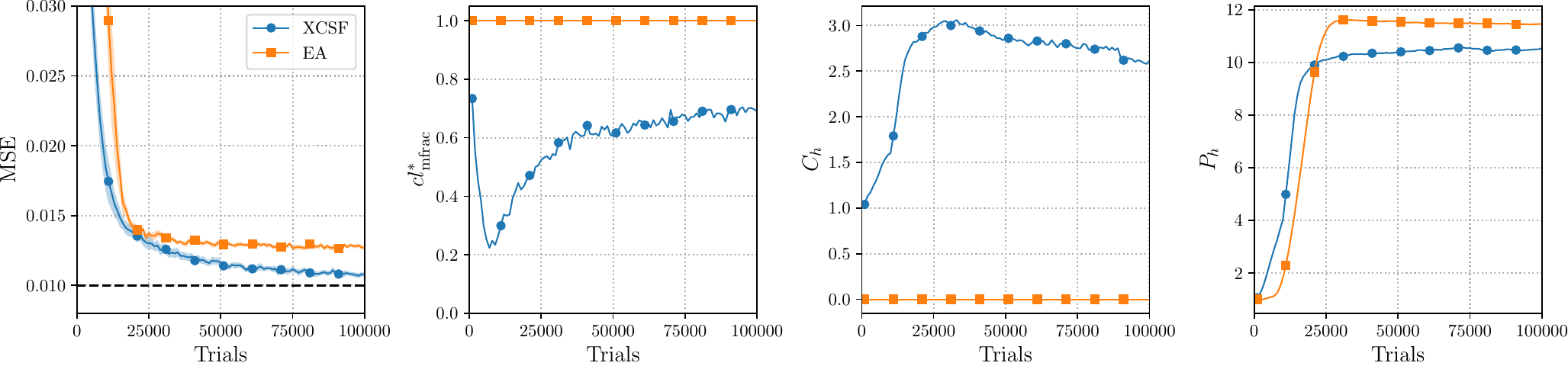}
    \caption{The performance of the EA and XCSF with maximum number of hidden neurons $h_{\text{max}}=12$ on \textsc{usps digits}. Shown are the mean squared error (MSE), fraction of inputs matched by the best rule ($cl^{*}_{\text{mfrac}}$), condition hidden neurons ($C_h$), and prediction hidden neurons ($P_h$), for the EA (square) and XCSF (circle). Connection mutation not applied; $h_M=1$.}
    \label{fig:12max}
\end{figure*}

The performance of XCSF and the EA on \textsc{usps digits} when the maximum number of hidden neurons $h_{\text{max}}=12$ is fixed below that which $\epsilon_0$ can be attained with a global model is shown in Fig.~\ref{fig:12max} and Table~\ref{table:errors}; connection mutation is disabled and $h_M=1$. After 100000 trials, XCSF attains a significantly smaller training error than the EA. The XCSF AUC = 1374.65 and EA = 1700.94, confirming that XCSF is able to partition the input space and achieve a smaller error than possible with a global solution. 

Applying the evolved autoencoders to a simple denoising task shows that this improvement in XCSF training error translates to a smaller test error. After adding 10\% salt and pepper noise to the test set inputs and comparing the reconstructions with the original clean inputs, the XCSF MSE ($0.017\pm0.001$) is significantly smaller than the EA ($0.021\pm0.001$).

\subsection{Summary}

Across all data sets, XCSF provides faster convergence over the first 100000 trials than the EA (AUC metric). Furthermore, across all data sets, XCSF can be seen to achieve faster early convergence, and for finding a global solution is always at least as fast as the EA in number of trials to $\epsilon_0$. While connection mutation was found to slow convergence on the simple \textsc{usps digits} data set, it was found beneficial to the search process on \textsc{cifar10}, which contains a large number of inputs and highly correlated features (i.e., RGB channels). 

Moreover, when the number of hidden neurons is restricted below which a global solution can reach the target error, XCSF has been shown capable of subdividing the input domain to achieve a smaller reconstruction error than is possible with the EA. For a given maximum allowable distortion $\epsilon_0$, the resulting decoders can therefore be made smaller and require less computation to perform decoding, which may be beneficial for low powered devices, etc.

Grouping the mean errors at 100000 trials for all data sets by algorithm, and applying the Wilcoxon signed-rank test shows that XCSF has a significantly smaller error than the EA, $p\le0.0077$. A summary of the discovered autoencoding architectures can be seen in Table~\ref{table:summary}.

\begin{table}[t]
    \centering
	\caption{Summary of discovered autoencoding architectures and time to $\epsilon_0$ after 100000 trials. Mean values reported.}
    \begin{tabular}{lrrrrr}
        \toprule
        \multicolumn{1}{l}{Algorithm} & 
        \multicolumn{1}{c}{$C_h$} &
        \multicolumn{1}{c}{$P_h$} &
        \multicolumn{1}{c}{$C_w$} &
        \multicolumn{1}{c}{$P_w$} &
        \multicolumn{1}{c}{$\epsilon_0$ Trials}\\
        \midrule
        \underline{\textsc{mnist digits}} &&&&&\\
        XCSF $h_M$=1 & 2.1 & 30.3 & 1662 & 47496 & 97000 \\
        EA $h_M$=1 & n/a & 27.6 & n/a & 43274 & n/a \\
        XCSF $h_M$=2 & 2.1 & 41.5 & 1651 & 65110 & 43000 \\
        EA $h_M$=2 & n/a & 40.7 & n/a & 63801 & 47000 \\
        \\
        \underline{\textsc{mnist fashion}} &&&&&\\
        XCSF $h_M$=1 & 2.7 & 18.6 & 2150 & 29201 & n/a \\
        EA $h_M$=1 & n/a & 19.3 & n/a & 30294 & n/a \\
        XCSF $h_M$=5 & 10.8 & 68.7 & 8455 & 107671 & 55000 \\
        EA $h_M$=5 & n/a & 64.2 & n/a & 100667 & 64000 \\
        \\
        \underline{\textsc{usps digits}} &&&&&\\
        XCSF $h_M$=1 & 2.1 & 16.8 & 534 & 8593 & 16000 \\
        EA $h_M$=1 & n/a & 17.4 & n/a & 8940 & 19000 \\
        XCSF-cmut $h_M$=1 & 1.9 & 18.6 & 242 & 6689 & 33000 \\
        EA-cmut $h_M$=1 & n/a & 19.0 & n/a & 7008 & 28000 \\
        \\
        \underline{\textsc{cifar10}} &&&&&\\
        XCSF $h_M$=5 & 7.6 & 50.9 & 23374 & 312956 & n/a \\
        EA $h_M$=5 & n/a & 41.4 & n/a & 253995 & n/a \\
        XCSF-cmut $h_M$=5 & 7.2 & 67.3 & 11245 & 276113 & 100000 \\
        EA-cmut $h_M$=5 & n/a & 64.3 & n/a & 265566 & n/a \\
        \bottomrule
    \end{tabular}
    \label{table:summary}
\end{table}

For an identical number of trials, XCSF and the EA are presented with the same number of training samples. However, it should be noted that forward and backward propagation of $cl.P$ networks are performed for all members of $[P]$ with the EA, whereas only those in $[M]$ are propagated with XCSF. In this regard, XCSF performs conditional computation and can be seen as a form of dynamic pruning based on gating functions~\cite{Han:2021}, as well as sharing characteristics with sparse autoencoders~\cite{Gong:2015}.

\section{Conclusion}
\label{section:conclusion}

Autoencoding is a key component of many learning systems and this article has presented the first results from using a variant of XCSF to perform such dimensionality reduction. The traditional approach to autoencoding involves manually specifying the number of neurons and using predefined constraints. The approach outlined here automatically identifies the minimal number of neurons required to reach a target error $\epsilon_0$---under this threshold, the system focuses on increasing the generality of solutions and pruning neurons. 

LCS enable the emergence of an ensemble of structurally heterogeneous solutions to cover the problem space. In this case, when the number of neurons in the autoencoders is allowed to evolve, networks of differing complexity are typically seen to cover different areas of the problem space. Furthermore, the scheme introduced here entirely self-adapts the search process: both the gradient-free mutation of weights and their local refinement where gradient information is available. Not only does this potentially reduce the number of hand-tunable parameters, it may provide further benefits in network analysis, use in non-stationary and online domains, etc. Moreover, the LCS ensemble may reveal input categories more clearly than are seen in a global network solution. Given their basis in EAs, LCS do not require the existence of helpful gradients within the weight space, although gradient-based search can speed learning, as here.

XCSF adaptively subdivides the input domain into local approximations that are simpler than a global neural network solution, enabling a more efficient allocation of reward. Given this difference in credit allocation, further improvement in convergence might result from certain architectural modifications to the classifiers. For instance, in the present system, both condition and prediction networks use a similar architecture and are subject to the same offspring creation algorithm. It seems possible that using separate algorithms allowing wider relative complexities could reveal higher rates of convergence.

Current work is therefore exploring additional layers of autoencoding and alternative layer architectures such as those involving local connectivity and weight sharing (e.g., convolutional layers) as well as those with recurrent connections (e.g., long short-term memory layers). A future examination of alternative classifier condition representations, e.g., symbolic trees or Haar-like features, and enhancements to the evolutionary operators may yield further improvements for specific applications. Future application to supervised and reinforcement learning tasks, e.g., under a pretraining scheme, will determine the effectiveness of the suggested dimensionality reduction shown here on various data sets.

%\bibliographystyle{IEEEtran}
%\bibliography{abrv,references}


\begin{IEEEbiography}[{\includegraphics[width=1in,height=1.25in,clip,keepaspectratio]{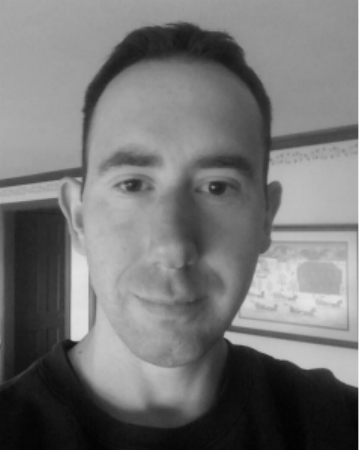}}]{Richard~J.~Preen} received the B.Sc.\ (Hons.) and M.Sc.\ degrees in computer science, and the Ph.D.\ degree in artificial intelligence from the University of the West of England, Bristol, U.K., in 2004, 2008, and 2011, respectively.

He is currently a Research Fellow with the Department of Computer Science and Creative Technologies at the University of the West of England.
\end{IEEEbiography}

\begin{IEEEbiography}[{\includegraphics[width=1in,height=1.25in,clip,keepaspectratio]{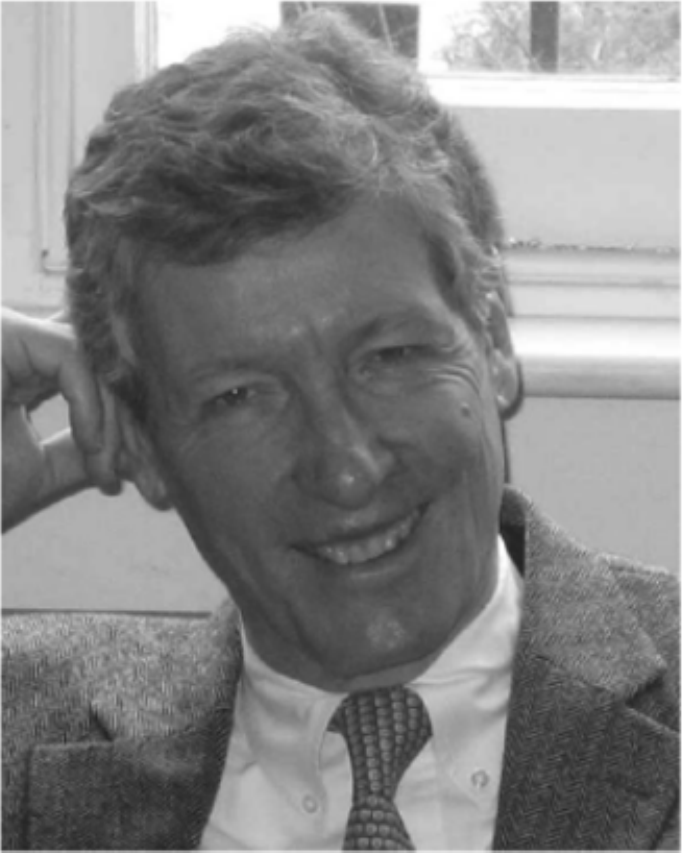}}]{Stewart~W.~Wilson} is an independent researcher in classifier systems and evolutionary computation. He received his S.B.\ (physics) and S.M.\ and Ph.D.\ (electrical engineering) from the Massachusetts Institute of Technology. His Ph.D.\ thesis under Edwin Land investigated what would happen if a child could ask questions and receive excellent spoken answers from a machine that connected him to recordings made by a scientist such as Carl Sagan. The results showed long, thoughtful lines of inquiry for this student-driven style of learning. Later, at Polaroid Corporation, Wilson developed a first practical implementation.

In 1981 he became interested in John Holland's classifier system idea, contacted Holland and his incipient group, and took up this machine learning approach. The results, eventually, were the algorithms XCS and XCSF which made Holland's idea practical and stimulated applications and further research worldwide. From his interest in vision, Wilson did research that connected the stroboscopic imagery of Gray Walter to the structure of the retino-cortical mapping, suggesting the latter contains a scanning mechanism.  In recent years, Wilson developed a system based on genetic algorithms for automatic investment portfolio definition. The concept found application  with SumGrowth Strategies, LLC in its implementation of Temporal Portfolio Theory.

Dr.~Wilson is a co-founder of \emph{Adaptive Behavior} and the Simulation of Adaptive Behavior (SAB) Conferences.
\end{IEEEbiography}

\vfill\eject

\begin{IEEEbiography}[{\includegraphics[width=1in,height=1.25in,clip,keepaspectratio]{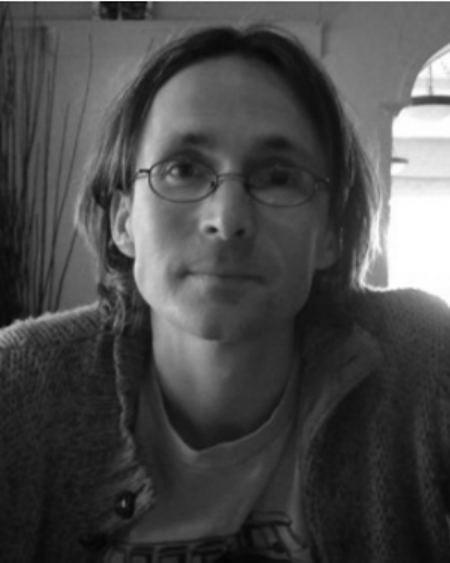}}]{Larry~Bull} received both the B.Sc.\ (Hons.) degree in computing for real-time systems and the Ph.D.\ degree from the University of the West of England, Bristol, U.K., in 1992 and 1995, respectively.

He is a Professor of artificial intelligence with the Department of Computer Science and Creative Technologies, University of the West of England. Prof.~Bull was the founding Editor-in-Chief of the journal \emph{Evolutionary Intelligence} and has published widely on nature inspired computation, including the monograph \emph{The Evolution of Complexity} (Springer, 2020).
\end{IEEEbiography}

\end{document}